\documentclass[runningheads]{llncs}
\usepackage{algorithm}
\usepackage[section]{placeins}
\usepackage[noend]{algpseudocode}
\usepackage{tabularx}
\usepackage{graphicx}
\usepackage{caption}    
\usepackage{amsmath,amssymb}
\usepackage{color}
\usepackage{mathtools}
\usepackage{diagbox}
\usepackage{array}
\usepackage{soul}
\graphicspath{{./figures/}}
\usepackage{graphicx}
\usepackage{hyperref}
\usepackage{multirow}
\usepackage{tikz}
\usepackage[caption=false]{subfig}
\usepackage{pgfplots}
\usetikzlibrary{pgfplots.groupplots}
\usetikzlibrary{matrix}
\usepackage[misc,geometry]{ifsym} 

\pgfplotsset{compat=newest}
\newcolumntype{P}[1]{>{\centering\arraybackslash}p{#1}}
\begin{document}

\usepgfplotslibrary{groupplots}
\title{Layered Embeddings for Amodal Instance Segmentation}
%
%
\author{Yanfeng Liu\textsuperscript{\Letter}\orcidID{0000-0001-8865-1972} \and
Eric T. Psota\orcidID{0000-0002-7836-298X} \and
Lance C. P\'{e}rez\orcidID{0000-0002-9733-9347}}
\authorrunning{Y. Liu et al.}
%
\institute{University of Nebraska-Lincoln, Lincoln, Nebraska, USA \\
\email{\{yanfeng.liu, epsota, lperez\}@unl.edu}}
\maketitle              
\begin{abstract} The proposed method extends upon the representational output of semantic instance segmentation by explicitly including both visible and occluded parts. A fully convolutional network is trained to produce consistent pixel-level embedding across two layers such that, when clustered, the results convey the full spatial extent and depth ordering of each instance. Results demonstrate that the network can accurately estimate complete masks in the presence of occlusion and outperform leading top-down bounding-box approaches. Source code available at \url{https://github.com/yanfengliu/layered\_embeddings}

\keywords{Semantic Instance Segmentation \and Amodal Segmentation \and Pixel Embedding \and Occlusion Recovery.}
\end{abstract}
\section{Introduction}
Instance segmentation extends semantic segmentation by distinguishing between objects of the same class. Instance segmentation methods assign a single instance label to visible pixels, thus each object's full spatial occupancy and depth ordering --- two properties that humans instinctively estimate --- are not represented in the output.
In contrast, when occluded regions are taken into consideration, this is referred to as \textit{amodal segmentation} \cite{li2016amodal}. 

To successfully segment occluded regions, the method not only needs to know where occlusions happen, but also the shape of unseen object parts relative to what is observed. If the objects are non-rigid, there can be multiple plausible solutions. On top of the inherent difficulty of the task, the lack of amodal ground truth makes it difficult to develop and evaluate new methods. 
Li and Malik \cite{li2016amodal} composited training data from PASCAL VOC 2012 \cite{Everingham10} by overlaying foreground masks. 
However, the resulting images are unnatural, with unrealistic lighting and object scales. 
Ehsani et al. \cite{ehsani2018segan} introduced a synthetic dataset ``DYCE" consisting of images rendered from various indoor graphics models at different angles. Zhu et al. \cite{zhu2017semantic} introduced the COCO Amodal dataset, consisting of thousands of amodal masks approximated by human annotators. They also provide baseline methods along with suggested performance metrics. Unfortunately, fundamental flaws in the data generation process (e.g., unrealistic renderings and inconsistent object labels and depth ordering), human errors (e.g., shadows inconsistently being labelled as instances of objects and synonyms/typos for classification labels), and an insufficient number of training images make it difficult to develop and analyze amodal segmentation methods.

We propose a fully-convolutional, end-to-end trainable approach that jointly estimates the presence of occlusion and provides consistent instance labeling across foreground and occluded regions. The method is evaluated on an easily configurable synthetic dataset consisting of various types of shapes with occlusions with precisely known amodal masks. Results demonstrate that the method is capable of accurately estimating layered spatial occupancy and outperforming a state-of-the-art top-down alternative. 

\section{Related Work} 
\label{sec2} 

Because ground truth instance labels are permutation invariant, the common approach of training deep fully-convolutional networks (FCNs) to detect and segment objects faces the dilemma of an ambiguous target \cite{long2015fully}. There are generally two categories of approaches used to achieve instance segmentation. Top-down methods begin by finding the regions (often bounding boxes) that contain each instance, and then performing pixel-wise segmentation of the dominant instance within that region. For example, Mask R-CNN \cite{he2017mask} extends Faster R-CNN \cite{ren2015faster} by adding a branch for segmentation mask prediction in parallel with the other branches (bounding boxes and classification). Li et al. \cite{li2017fully} proposed an alternative that uses location-sensitive fully convolutional networks to partition bounding boxes into 3$\times$3 grids, and then evaluates the likelihood that each partition contains the correct part relative to the other partitions. 

The second category, bottom-up methods, begin by assigning attributes to pixels and then clustering them into instances. Examples include those that use pixel embedding to move the high-level detection stage to the end of the process \cite{fathi2017semantic}\cite{de2017semantic}. Fathi et al., \cite{fathi2017semantic} adopt this principle by training a network to evaluate pairwise pixel similarity. They train a separate model to generate seed points that represent the typicality of a pixel compared to other pixels in the area. Brabandere et al., \cite{de2017semantic} proposes a discriminative loss function to train pixel embeddings such that they are close within the same instance but far apart for different instances.
                                                                      
The above methods propose a surjective mapping from pixels to instances. However, it is worth considering if this is an optimal representation of semantic instance segmentation. Computer vision often aims to reverse-engineer scenes from images/video, and an assignment of all visible parts to a single membership is an incomplete descriptor. In contrast, the full segmentation masks and relative depth ordering prior to image projection provides a more complete descriptor.

To this end, Yang et al. \cite{5540070} \cite{6042883} estimate layer ordering as part of instance segmentation and introduce a learned predictor based on relative detection scores, position on the ground plane, and size. They acknowledge the benefits of full spatial segmentations of visible and occluded parts, but their method focuses on the benefits of depth ordering for instance grouping. Chen et al. \cite{7298969} attempt to fill occluded regions by selecting similar non-occluded exemplar templates from a library; this improves instance segmentation of visible pixels. Uhrig et al. \cite{uhrig2016pixel} propose to consider explicit depth ordering estimation for instance segmentation. Their method exploits ground truth depth information, but it does not attempt to recover occluded segments. While each of these methods uses the concept of occlusions to improve instance segmentation, none of them explicitly targets the full spatial extents and depth ordering of instances.

Li and Malik \cite{li2016amodal} use an iterative approach to gradually predict the amodal masks based on the bounding box and classification produced by an object detector. They compute the visible mask and iteratively expand upon it to produce the amodal mask and bounding box. Ehsani et al. \cite{ehsani2018segan} propose a GAN-based model to produce both the segmentation and the appearance of the occluded regions, assuming that foreground segmentation is already pre-calculated by other methods. However, their method focuses on crops with one salient object.

Amodal segmentation remains as a challenging task and very few studies and datasets exist. Current methods either focus on a special case of the general problem or extend upon top-down approaches. This paper proposes an alternative bottom-up approach and examines some challenges associated with amodal segmentation.

\section{Method}

The goal of the proposed method is to produce full instance masks for each segmented object as long as part of the object is visible in the image. To circumvent the limitations of DYCE and COCO Amodal datasets, a synthetically generated dataset of shapes is used. The advantages of this set include 1) full control of scene complexity; 2) access to precise ground truth; and 3) rigid shapes where the ground truth is often unique given partial observations. 

The dataset has three classes of shapes: triangles, rectangles, and circles. All shapes have a fixed size, but their locations, orientations, and depth orderings are randomized. Shapes have the same color as the background, only distinguished by their black outlines, so that the network cannot cheat by simply detecting color or intensity. This representation forces the network to rely on outlines and be aware of large regions for context. 

\begin{figure}[t!]
\centering
\begin{tabular}{ccccc}
\includegraphics[width=0.17\textwidth, keepaspectratio]{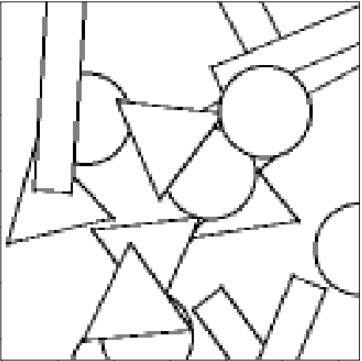}&
\includegraphics[width=0.17\textwidth, keepaspectratio]{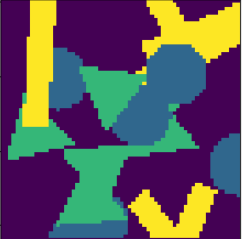}&
\includegraphics[width=0.17\textwidth, keepaspectratio]{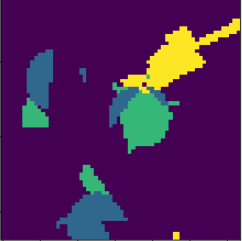}&
\includegraphics[width=0.17\textwidth, keepaspectratio]{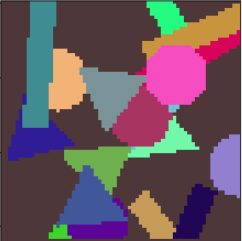}&
\includegraphics[width=0.17\textwidth, keepaspectratio]{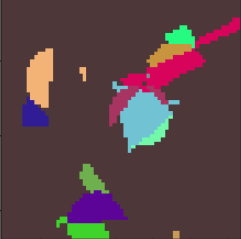}\\
\end{tabular}
\caption{Sample training data. From left to right: input image, foreground class mask, occlusion class mask, foreground instance mask, occlusion instance mask}
\label{sample_training}
\end{figure}

For training and evaluation, the ground truth masks are arranged in the following manner: foreground semantic classification, occlusion semantic classification, foreground instance labels, and occlusion instance labels. See Fig. \ref{sample_training} for a sample of training data. The proposed method generates the following four outputs: 1) \textit{foreground multi-class semantic segmentation}, that labels pixels as background, foreground, or occluded; 2) \textit{occlusion multi-class semantic segmentation}, that labels occluded pixels as one of the classes; 3) \textit{foreground embedding}, used to cluster foreground pixels into instances; 4) \textit{occlusion embeddings}, that are consistent with visible instances. It is worth noting that occluded pixels are defined as those where one instance occludes another instance. While the method does not consider occlusions caused by background objects, this could trivially be added as another class to the output.

To train instance embedding across both the foreground and occluded regions, the method uses a variation of the discriminative loss function introduced in \cite{de2017semantic} . 
Consider an input image $\mathcal{I}$ and a pair of embedding outputs $\mathcal{E}_f$ and $\mathcal{E}_o$ that contain embedding vectors for the foreground region and occluded region, respectively. 
The embedding outputs are matrices with the same spatial dimensions (rows and columns) as the input image, where the number of channels $C$ represents the dimensionality of each pixel's embedding. The goal of the network is to map each foreground pixel $p \in \mathcal{I}$ to a $C$-dimensional embedding vector $\mathcal{E}_f(p)$ and each occluded pixel $p \in \mathcal{I}$ to a $C$-dimensional embedding vector $\mathcal{E}_o(p)$ such that embedding vectors for pixels belonging to the same instance are close together in the $C$-dimensional space and embedding vectors for different instances are far apart.

\begin{figure}[t!]
\centering
\includegraphics[width = 1.0\textwidth]{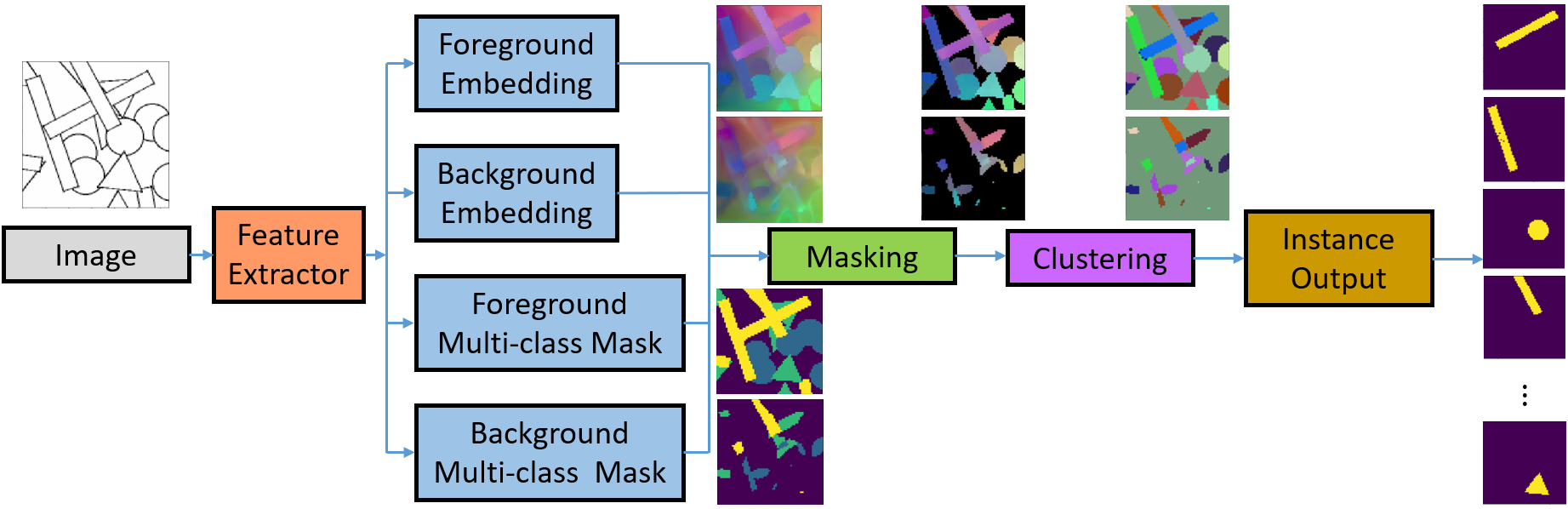}
\vspace{-0.5cm}
\caption{Proposed segmentation architecture.}
\label{sample_output}
\end{figure}

The overall loss consists of three terms: variance $l_{var}$, distance $l_{dst}$, and regularization $l_{reg}$. Let $K$ be the total number of classes, $N_k$ be the number of instances of class $k$, $N$ be the number of ground truth instances and let $\mathcal{R}_f^n \subseteq \mathcal{I}$ and $\mathcal{R}_o^n\subseteq \mathcal{I}$ denote the set of foreground and occluded pixels for instance $n \in \{1,\ldots,N\}$, respectively. Also, let $\mu_n$ be the average embedding vector of all pixels in both the foreground and occlusion embeddings for instance $n$. The variance term and distance term are defined as

\begin{equation}
\resizebox{\hsize}{!}{$l_{var} = \frac{1}{N}\sum^N_{n=1}\frac{1}{|\mathcal{R}_f^n| + |\mathcal{R}_o^n|} \left(\sum_{p \in \mathcal{R}_f^n}\big[||\mu_n-\mathcal{E}_f(p)||-d_{var}\big]_{+}^2 + \sum_{p \in \mathcal{R}_o(n)}\big[||\mu_n-\mathcal{E}_o(p)||-d_{var}\big]_{+}^2 \right)$}
\end{equation}
and
\begin{equation}
l_{dst} = \sum_{k=1}^K\frac{1}{N_k(N_k-1)}{\sum^{N_k}_{n = 1}\sum^{N_k}_{\substack{m=1 \\ m\neq n}}}\big[2d_{dst} - ||\mu_{n} - \mu_{m}||\big]_{+}^2,
\end{equation}
where $[a]_{+} = max(a, 0)$ is the hinge loss, and $||\cdot||$ is L1 distance. 
Constants $d_{var}$ and $d_{dst}$ are the margins for the variance and distance term.
Effectively, $l_{var}$ penalizes pixels that belong to the same instance but are farther than $d_{var}$ apart in the embedding space, and $l_{dst}$ penalizes cluster centers that represent different instances but are closer than $d_{dst}$.
The regularization term 
\begin{equation}
l_{reg} = \frac{1}{N}\sum^N_{n=1}||\mu_n||
\end{equation}
prevents the network from minimizing $l_{dst}$ by simple embedding amplification.
Finally, the network is trained to minimize $L_{total} = \alpha \cdot l_{var} + \beta \cdot l_{dst} + \gamma \cdot l_{reg}$. 
Fig. \ref{sample_output} presents an example of the method applied to shapes.

The network uses a pre-trained feature extractor and produces a depth-concatenation of four outputs. The four modules share the output from the feature extractor. The method clusters the network's pixel embeddings into instances using the algorithm presented in \cite{de2017semantic}. A random unlabeled pixel is selected and the embeddings around it within $v_{var}$ are grouped together. The mean embedding of this group is used for the next round of grouping until convergence. 

\section{Implementation Details}
The proposed method uses DeeplabV3+ \cite{chen2018encoder} with an Xception backbone as the feature extractor. Its final upsampling and logit layers are removed and the 256-dimensional output is used as features. Input size is set to 256x256 and the output size is 64x64. For the loss function, $\alpha = \beta = \gamma = 1, d_{var} = 0.5, d_{dst} = 1.5$. Embedding dimension $C = 6$ and the mean shift threshold for clustering is 1.5, which is consistent with $d_{dst}$. The embedding module consists of 256, 256, 128, and $C$ convolution filters, with RELU activations. 

For comparison, Mask R-CNN is selected as a representative model of top-down approaches for baseline due to the success and popularity of the R-CNN family among object detection and segmentation architectures. It is an architecture originally designed only for foreground instance segmentation, but it can be easily modified to perform amodal instance segmentation by fine-tuning on amodal ground truth. Its weights are pre-trained on MS COCO. Its output is upsampled from the original $m \times m$ mask to the corresponding bounding box size and put in the context of the whole image. The final output is resized to 64x64 to be directly comparable with our method.

All models are trained for 100 epochs in Tensorflow with Adam optimization (learning rate = 0.0001 and batch size = 2). Training examples are generated at runtime with the same initial random seed. Each epoch has 1000 images.

\section{Results}
The embedding model and Mask R-CNN are both evaluated on 1000 shapes images randomly generated with 6 instances per class. Results in Table \ref{instance_layer_gt} show that our model outperforms Mask R-CNN on AP, AP\textsubscript{50}, AR\textsubscript{None}, and AR\textsubscript{Partial}. Mask R-CNN achieves better results on AP\textsubscript{75} when non-max suppression threshold $t=0.7$, and AR\textsubscript{100}, AR\textsubscript{Heavy} when $t = 0.9$, but it cannot retain high performance with a single value of $t$. The same performance pattern repeats when the number of instances per class is increased from 6 to 12.

\begin{table*}[b!]
\centering
\caption{Performance of models on 6 instances per class with different NMS threshold $t$ and different number of instances per class $N$}
\label{instance_layer_gt}
\begin{tabularx}{\textwidth}{|l|l|X|X|X|X|X|X|X|l|}
\hline
$N$ & Model & AP & AP\textsubscript{50} & AP\textsubscript{75} & AR\textsubscript{100} & AP\textsubscript{None} & AR\textsubscript{Partial} & AP\textsubscript{Heavy}\\
\hline
\multirow{6}{*}{6} & Ours & \bfseries 0.7673 & \bfseries 0.9091 & 0.7800 & 0.7933 & \bfseries 0.9983 & \bfseries 0.9637 & 0.6190 \\ 
 \cline{2-9}
 & $\text{MRCNN}_{t=0.1}$ & 0.5553 & 0.6526 & 0.6249 & 0.5823 & 0.7916 & 0.6986 & 0.4373 \\ 
 \cline{2-9}
 & $\text{MRCNN}_{t=0.3}$ & 0.6781 & 0.8010 & 0.7706 & 0.7132 & 0.8765 & 0.8417 & 0.5898 \\ 
 \cline{2-9}
 & $\text{MRCNN}_{t=0.5}$ & 0.7238 & 0.8701 & 0.8187 & 0.7620 & 0.9017 & 0.8768 & 0.6562 \\ 
 \cline{2-9}
 & $\text{MRCNN}_{t=0.7}$ & 0.7332 & 0.8860 & \bfseries 0.8323 & 0.7766 & 0.9112 & 0.8822 & 0.6748 \\ 
 \cline{2-9}
 & $\text{MRCNN}_{t=0.9}$ & 0.6109 & 0.7334 & 0.6800 & \bfseries 0.8010 & 0.9240 & 0.8978 & \bfseries 0.6998 \\ 
 \hline \hline
\multirow{6}{*}{12} & Ours & \bfseries 0.5262 & \bfseries 0.7099 & 0.5192 & 0.5697 & \bfseries 0.9958 & \bfseries 0.9499 & 0.4049 \\ 
 \cline{2-9}
 & $\text{MRCNN}_{t=0.1}$ & 0.3564 & 0.4350 & 0.4068 & 0.3730 & 0.7247 & 0.6220 & 0.2562 \\ 
 \cline{2-9}
 & $\text{MRCNN}_{t=0.3}$ & 0.4641 & 0.5827 & 0.5312 & 0.4887 & 0.8226 & 0.7678 & 0.3715 \\ 
 \cline{2-9}
 & $\text{MRCNN}_{t=0.5}$ & 0.5127 & 0.6416 & 0.5874 & 0.5399 & 0.8618 & 0.8019 & 0.4290 \\ 
 \cline{2-9}
 & $\text{MRCNN}_{t=0.7}$ & 0.5203 & 0.6570 & \bfseries 0.5926 & 0.5533 & 0.8722 & 0.8110 & 0.4424 \\ 
 \cline{2-9}
 & $\text{MRCNN}_{t=0.9}$ & 0.4022 & 0.5253 & 0.4450 & \bfseries 0.5856 & 0.8913 & 0.8326 & \bfseries 0.4702 \\ 
 \hline
\end{tabularx}
\end{table*}

As part of an ablation study, the semantic segmentation prediction for the model is replaced with ground truth. With 6 instances per class, AP increases from 0.7673 to 0.7959 and AR\textsubscript{100} increases from 0.7933 to 0.8300 when using ground truth; with 12 instances per class, AP increases from 0.5262 to 0.5419 and AR\textsubscript{100} increases from 0.5697 to 0.6013. These marginal improvements suggest that semantic segmentation is not the primary bottleneck of performance.

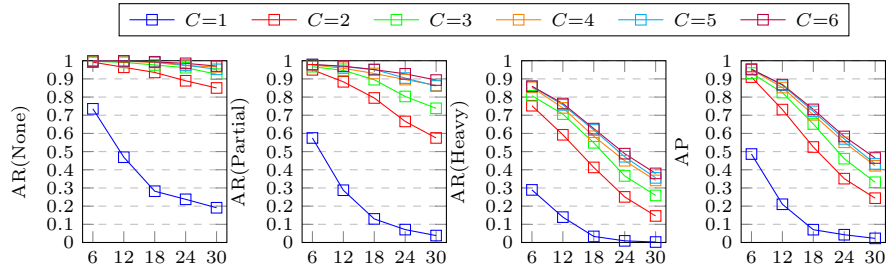
\begin{figure}[t!]
\centering
\begin{tikzpicture}
\tikzstyle{every node}=[font=\scriptsize]
\begin{groupplot}[group style={
    group name = metrics,
    group size= 4 by 1, horizontal sep=1cm, vertical sep=1cm},height=4cm,width=3.5cm]
\nextgroupplot[
    ylabel={AR(None)},
    xmin=4, xmax=32,
    ymin=0, ymax=1,
    xtick={6, 12, 18, 24, 30},
    ytick={0, 0.1, 0.2, 0.3, 0.4, 0.5, 0.6, 0.7, 0.8, 0.9, 1.0},
    legend pos=north east,
    ymajorgrids=true,
    grid style=dashed,
]
\addplot[
    color=blue,
    mark=square,
    ]
    coordinates {
    (6, 0.7354)(12, 0.4692)(18, 0.2821)(24, 0.2375)(30, 0.1911)
    };\label{plots:plot1}

\addplot[
    color=red,
    mark=square,
    ]
    coordinates {
    (6, 0.9920)(12, 0.9638)(18, 0.9355)(24, 0.8896)(30, 0.8501)
    };\label{plots:plot2}

\addplot[
    color=green,
    mark=square,
    ]
    coordinates {
    (6, 0.9966)(12, 0.9909)(18, 0.9767)(24, 0.9615)(30, 0.9258)
    };\label{plots:plot3}

\addplot[
    color=orange,
    mark=square,
    ]
    coordinates {
    (6, 0.9986)(12, 0.9966)(18, 0.9904)(24, 0.9729)(30, 0.9662)
    };\label{plots:plot4}

\addplot[
    color=cyan,
    mark=square,
    ]
    coordinates {
    (6, 0.9994)(12, 0.9985)(18, 0.9944)(24, 0.9761)(30, 0.9526)
    };\label{plots:plot5}

\addplot[
    color=purple,
    mark=square,
    ]
    coordinates {
    (6, 0.9987)(12, 0.9989)(18, 0.9934)(24, 0.9856)(30, 0.9706)
    };\label{plots:plot6}
    
\nextgroupplot[
    ylabel={AR(Partial)},
    xmin=4, xmax=32,
    ymin=0, ymax=1,
    xtick={6, 12, 18, 24, 30},
    ytick={0, 0.1, 0.2, 0.3, 0.4, 0.5, 0.6, 0.7, 0.8, 0.9, 1.0},
    legend pos=north east,
    ymajorgrids=true,
    grid style=dashed,
]
\addplot[
    color=blue,
    mark=square,
    ]
    coordinates {
    (6, 0.5750)(12, 0.2875)(18, 0.1297)(24, 0.0709)(30, 0.0381)
    };

\addplot[
    color=red,
    mark=square,
    ]
    coordinates {
    (6, 0.9498)(12, 0.8840)(18, 0.7950)(24, 0.6660)(30, 0.5751)
    };

\addplot[
    color=green,
    mark=square,
    ]
    coordinates {
    (6, 0.9671)(12, 0.9469)(18, 0.8954)(24, 0.8034)(30, 0.7382)
    };

\addplot[
    color=orange,
    mark=square,
    ]
    coordinates {
    (6, 0.9771)(12, 0.9584)(18, 0.9299)(24, 0.8962)(30, 0.8642)
    };

\addplot[
    color=cyan,
    mark=square,
    ]
    coordinates {
    (6, 0.9794)(12, 0.9673)(18, 0.9519)(24, 0.9045)(30, 0.8610)
    };

\addplot[
    color=purple,
    mark=square,
    ]
    coordinates {
    (6, 0.9791)(12, 0.9702)(18, 0.9509)(24, 0.9279)(30, 0.8936)
    };

\nextgroupplot[
    ylabel={AR(Heavy)},
    xmin=4, xmax=32,
    ymin=0, ymax=1,
    xtick={6, 12, 18, 24, 30},
    ytick={0, 0.1, 0.2, 0.3, 0.4, 0.5, 0.6, 0.7, 0.8, 0.9, 1.0},
    ymajorgrids=true,
    grid style=dashed,
]
\addplot[
    color=blue,
    mark=square,
    ]
    coordinates {
    (6, 0.2893)(12, 0.1394)(18, 0.0334)(24, 0.0090)(30, 0.0025)
    };

\addplot[
    color=red,
    mark=square,
    ]
    coordinates {
    (6, 0.7524)(12, 0.5925)(18, 0.4130)(24, 0.2508)(30, 0.1459)
    };

\addplot[
    color=green,
    mark=square,
    ]
    coordinates {
    (6, 0.8074)(12, 0.7064)(18, 0.5484)(24, 0.3661)(30, 0.2590)
    };

\addplot[
    color=orange,
    mark=square,
    ]
    coordinates {
    (6, 0.8534)(12, 0.7404)(18, 0.5844)(24, 0.4497)(30, 0.3416)
    };

\addplot[
    color=cyan,
    mark=square,
    ]
    coordinates {
    (6, 0.8590)(12, 0.7570)(18, 0.6190)(24, 0.4717)(30, 0.3554)
    };

\addplot[
    color=purple,
    mark=square,
    ]
    coordinates {
    (6, 0.8595)(12, 0.7645)(18, 0.6266)(24, 0.4887)(30, 0.3801)
    };

\nextgroupplot[
    ylabel={AP},
    xmin=4, xmax=32,
    ymin=0, ymax=1,
    xtick={6, 12, 18, 24, 30},
    ytick={0, 0.1, 0.2, 0.3, 0.4, 0.5, 0.6, 0.7, 0.8, 0.9, 1.0},
    ymajorgrids=true,
    grid style=dashed,
]
\addplot[
    color=blue,
    mark=square,
    ]
    coordinates {
    (6, 0.4866)(12, 0.2103)(18, 0.0704)(24, 0.0427)(30, 0.0226)
    };

\addplot[
    color=red,
    mark=square,
    ]
    coordinates {
    (6, 0.9098)(12, 0.7311)(18, 0.5252)(24, 0.3511)(30, 0.2437)
    };

\addplot[
    color=green,
    mark=square,
    ]
    coordinates {
    (6, 0.9321)(12, 0.8247)(18, 0.6518)(24, 0.4611)(30, 0.3314)
    };

\addplot[
    color=orange,
    mark=square,
    ]
    coordinates {
    (6, 0.9512)(12, 0.8524)(18, 0.7023)(24, 0.5512)(30, 0.4220)
    };

\addplot[
    color=cyan,
    mark=square,
    ]
    coordinates {
    (6, 0.9545)(12, 0.8652)(18, 0.7215)(24, 0.5698)(30, 0.4322)
    };

\addplot[
    color=purple,
    mark=square,
    ]
    coordinates {
    (6, 0.9521)(12, 0.8689)(18, 0.7323)(24, 0.5834)(30, 0.4648)
    };
    \end{groupplot}
    ;
\path (metrics c1r1.north west|-current bounding box.north)--
      coordinate(legendpos)
      (metrics c4r1.north east|-current bounding box.north);
\matrix[
    matrix of nodes,
    anchor= south,
    draw,
    inner sep=0.2em,
    draw
  ]at([yshift=1ex]legendpos)
  {
    \ref{plots:plot1}& $C$=1&[5pt]
    \ref{plots:plot2}& $C$=2&[5pt]
    \ref{plots:plot3}& $C$=3&[5pt]
    \ref{plots:plot4}& $C$=4&[5pt]
    \ref{plots:plot5}& $C$=5&[5pt]
    \ref{plots:plot6}& $C$=6\\};
\end{tikzpicture}
\caption{Performance for different number of instances per class and different embedding dimensions $C$. x-axis is number of instances, y-axis is different metrics, and colored lines represent models with different $C$.}
\label{instance_num}
\end{figure}

In the second part of the ablation study, the instance clustering result is replaced by ground truth. This way, the only source of error is insufficient layers of masks to accommodate the full complexity of occlusions. As Table \ref{gt_layers} shows, the performance gains diminishing improvement as the model allows more layers to represent occlusion. Data suggests that three or more layers of mask output in the embedding model instead of just ``foreground and occlusion" could improve performance, depending on the complexity of the application. 

\begin{figure}[t!]
\centering
\begin{tabular}{P{\linewidth}}
\includegraphics[width=\textwidth, keepaspectratio]{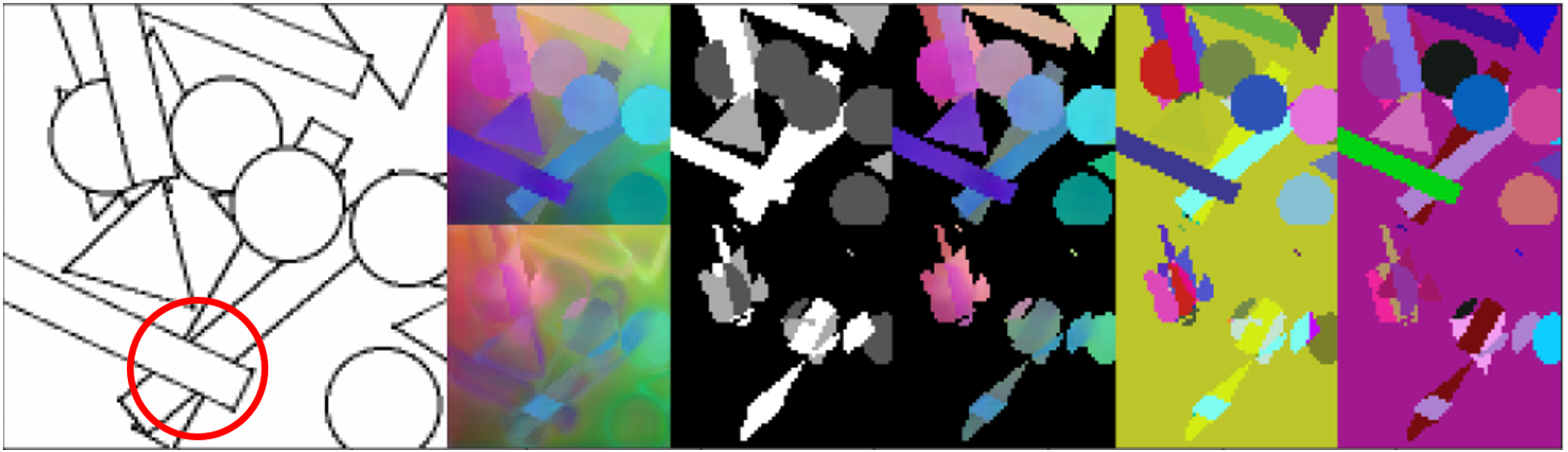}\\
(a) left to right: image, embedding, semantic segmentation mask, masked embedding, labels, ground truth. Top is foreground, and bottom is occlusion \\
\includegraphics[width=\textwidth, keepaspectratio]{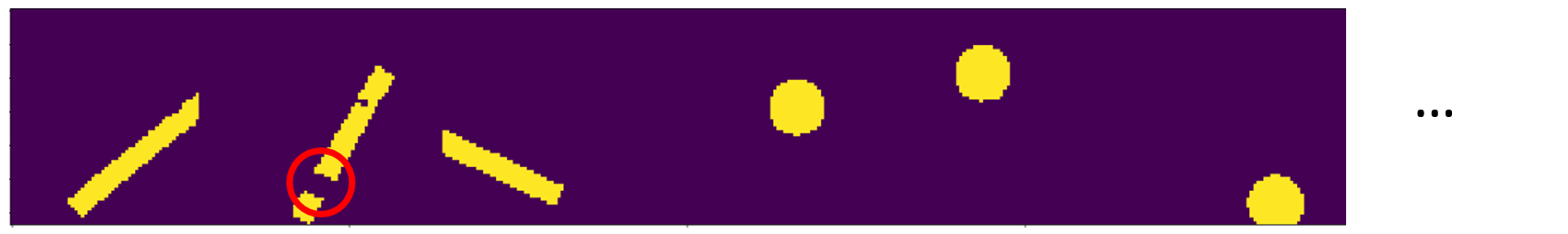}\\
(b) individual instance masks\\
\end{tabular}
\caption{Failure cases for embedding model. Red circle indicates a triple stack.}
\label{embedding_fail}
\end{figure}

The embedding model is also trained and evaluated on shapes datasets with different number of instances per class in order to study its embedding capacity. Since embeddings for shapes of different classes are allowed to be similar, the class is limited to rectangles in this study. Fig. \ref{instance_num} shows that the performance drops when there are more instances of the same class. This happens for two reasons: first, more instances introduce more occlusions, which increases the chance of more than two objects stacking together; second, distinguishing between more instances within the same class requires the model to have higher embedding capacity. Fig. \ref{instance_num} also shows that increasing the dimension results in diminishing but consistent improvement on performance. This is because the loss function encourages embeddings of the same instance to be close to one another, and embeddings of different instances to be far away. The penalty on the magnitude of embeddings makes this goal hard to achieve in low dimensions. In higher dimensions it is much easier to find another embedding that is both close to the origin and far enough from other embeddings.

The structure of the embedding model allows easier expansion into three or more layers of masks. Two layers shows its limits when objects are heavily occluded. For example, it is impossible to get correct results in Fig. \ref{embedding_fail} because there are three objects stacked within the red circle. The model can recover two of them at best. This is the most typical failure case for the proposed embedding model. By evaluating masks constructed from different number of layers of ground truth, Table \ref{instance_layer_gt} shows that more layers of masks will lead to better results and that when the number of layers is held constant, performance on heavily occluded regions gets worse because more instances are stacked.

\begin{figure}[t!]
\centering
\begin{tabular}{ccc|ccc}
\includegraphics[width=0.15\textwidth, keepaspectratio]{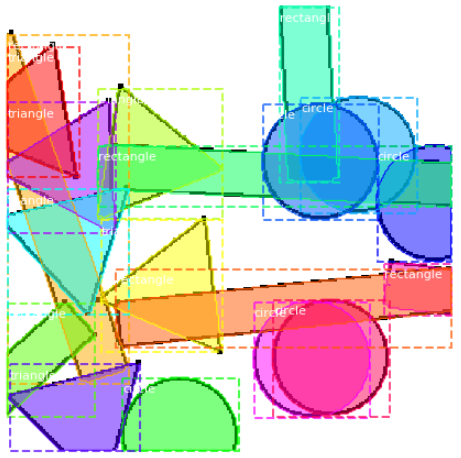}&
\includegraphics[width=0.15\textwidth, keepaspectratio]{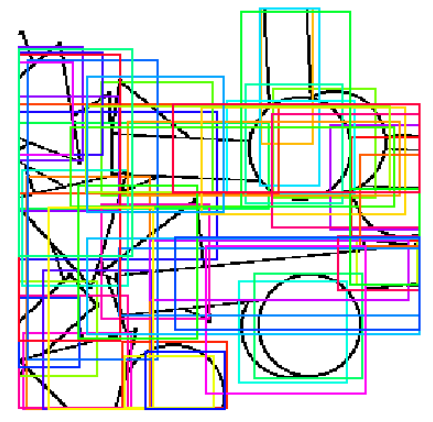}&
\includegraphics[width=0.15\textwidth, keepaspectratio]{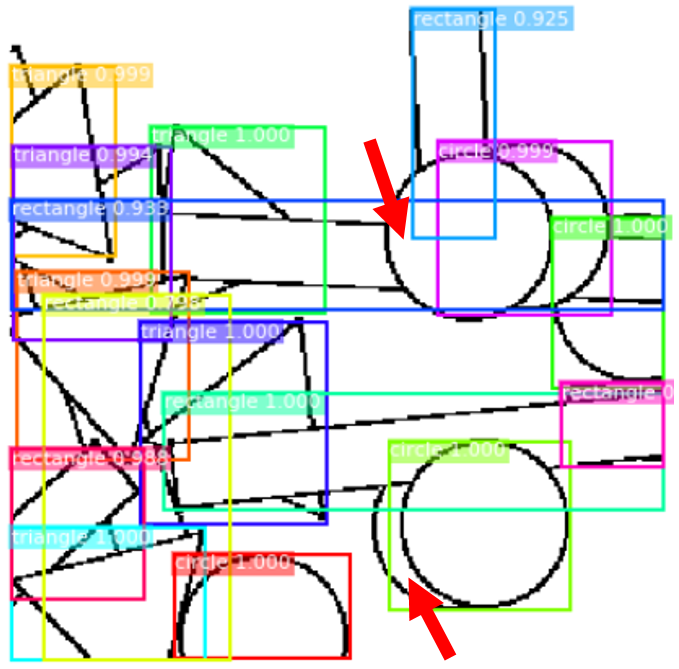}&
\includegraphics[width=0.15\textwidth, keepaspectratio]{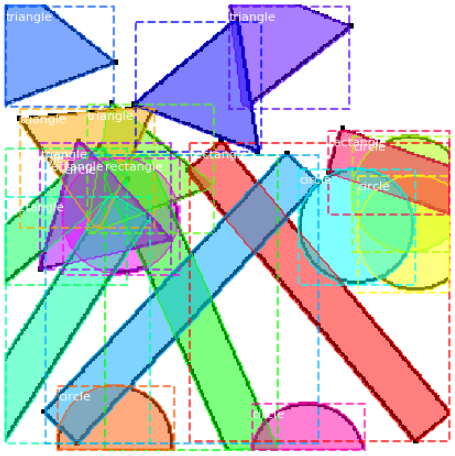}&
\includegraphics[width=0.15\textwidth, keepaspectratio]{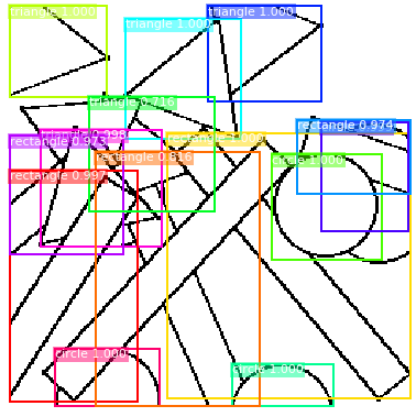}&
\includegraphics[width=0.15\textwidth, keepaspectratio]{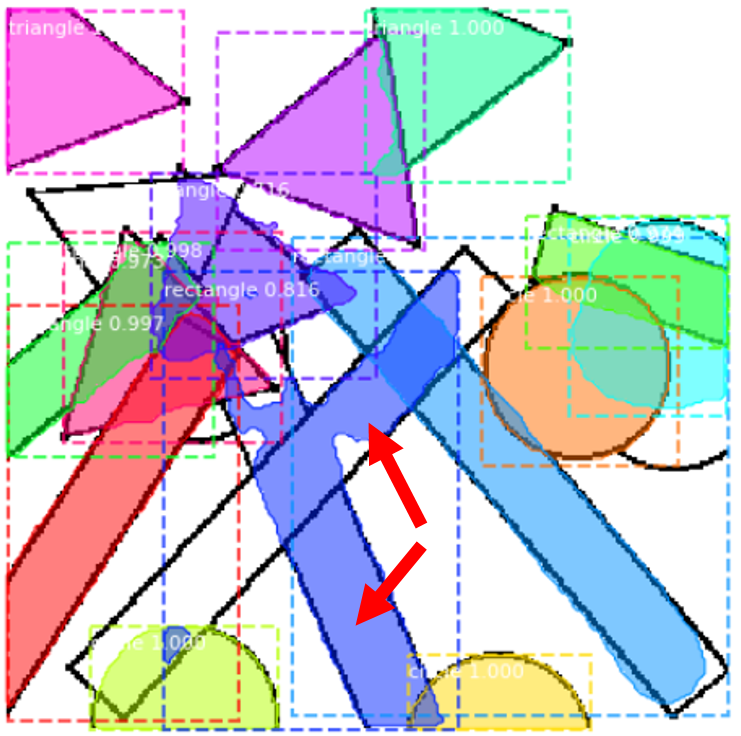}\\
\end{tabular}
\caption{Two failure cases for Mask R-CNN. From left to right, first case: ground truth, unfiltered proposals, filtered proposals; second case: ground truth, filtered proposals, final results}
\label{mrcnn_fail}
\end{figure}

\begin {table*}[t]
\centering
\caption{Performance of instance masks constructed from different number of layers of ground truth. $N$ is the number of instances per class.}
\label{gt_layers}
\begin{tabularx}{\textwidth}{|l|l|X|X|X|X|X|X|l|}
\hline
$N$ & Layers & AP & AP\textsubscript{50} & AP\textsubscript{75} & AR\textsubscript{100} & AP\textsubscript{None} & AR\textsubscript{Partial} & AP\textsubscript{Heavy}\\
\hline
\multirow{3}{*}{6} & 2 & 0.8584 & 0.9604 & 0.8614 & 0.8627 & 1.0000 & 0.9927 & 0.7408 \\ 
 \cline{2-9}
 & 3 & 0.9703 & 0.9901 & 0.9802 & 0.9744 & 1.0000 & 0.9997 & 0.9509 \\
 \hline \hline
\multirow{4}{*}{12} & 2 & 0.6594 & 0.8317 & 0.6436 & 0.6611 & 1.0000 & 0.9910 & 0.5251 \\
 \cline{2-9}
 & 3 & 0.8703 & 0.9703 & 0.8812 & 0.8729 & 1.0000 & 0.9993 & 0.8206 \\ 
 \cline{2-9}
 & 4 & 0.9584 & 0.9901 & 0.9703 & 0.9629 & 1.0000 & 0.9999 & 0.9475 \\ 
 \hline
\end{tabularx}
\end{table*}

Mask R-CNN has two typical failure cases, as shown in Fig. \ref{mrcnn_fail}. In the first case, the region proposal network generates the correct bounding boxes for the indicated circles, but some get filtered out during the non-maximum suppression stage. In the second case, both indicated rectangles fit within one bounding box and the mask generator is confused about which one is the salient object. 

This points to a fundamental difference between the two types of approaches. Top-down, bounding-box-based approaches have the ability to look at a region multiple times and potentially generate a complete mask each time. However, this also acts as a double-edged sword when multiple objects could appear in the same bounding box and compete for the mask if they are the same class. The embedding model, on the other hand, is a bottom-up, bounding-box-free approach. Each pixel can only have one final embedding per layer, which will then be clustered into an instance label or ignored as background. The trade-off is that, situations like Fig. \ref{mrcnn_fail} can be avoided, but, when number of layers of ground truth mask in the training data is insufficient, the best possible performance will have a relatively low upper bound. 

\section{Conclusion}
The method presented in this paper pushes the boundaries of a deep network's understanding of images by training it to estimate segmentation masks for unseen parts, and to associate them with visible instances. Experiments show that this bottom-up approach outperforms Mask R-CNN, a typical architecture for instance segmentation, by addressing the fundamental flaw in top-down approaches: the inability of bounding boxes to precisely capture the spatial relationship between cluttered objects. While the method only considers two-layer occlusion scenarios, the network structure can be easily modified to handle an arbitrary number of object types arranged in three or more layers. 

Because it is difficult to obtain accurate annotations of occluded parts, the proposed method instead uses a synthetic dataset for training. We hope this and other recent works will motivate the creation of large datasets with ground truth masks representing the full spatial occupancy of occluded instances. 

\bibliographystyle{splncs04}
\bibliography{egbib}
\end{document}